\documentclass[final]{cvpr}

\usepackage{times}
\usepackage{epsfig}
\usepackage{graphicx}
\usepackage{amsmath}
\usepackage{amssymb}
\usepackage{booktabs}

\usepackage{bm}
\usepackage{subcaption}
\usepackage{bbm}
\usepackage{makecell}
\usepackage{multirow}
\usepackage{cuted}

\usepackage[font={small}]{caption}

\usepackage[pagebackref=true,breaklinks=true,colorlinks,bookmarks=false]{hyperref}

\usepackage[misc]{ifsym}

\def\cvprPaperID{0822} 
\def\confYear{CVPR 2021}

\begin{document}

\title{Audio-Driven Emotional Video Portraits}


\author{Xinya Ji$^{1}$ \quad Hang Zhou$^{2}$\quad Kaisiyuan Wang$^3$ \quad Wayne Wu$^{4,5}$\thanks{Corresponding authors.} \quad Chen Change Loy$^{5}$ \\ Xun Cao$^{1}$\footnotemark[1] \quad Feng Xu$^{6}$\footnotemark[1]\\
    
    $^1$Nanjing University, \quad $^2$The Chinese University of Hong Kong, \\  $^3$The University of Sydney, \quad $^4$SenseTime Research, \\  
      $^5$S-Lab, Nanyang Technological University, \quad  $^6$BNRist and school of software, Tsinghua University\\
      {\tt\small  \{xinya@smail., caoxun@\}nju.edu.cn, zhouhang@link.cuhk.edu.hk, ccloy@ntu.edu.sg}, \\{\tt\small
       kaisiyuan.wang@sydney.edu.au, wuwenyan@sensetime.com, xufeng2003@gmail.com}    \vspace{-35pt}
}

\maketitle
\pagestyle{empty}
\thispagestyle{empty}

\begin{strip}
    \centering
    \includegraphics[width=1\textwidth]{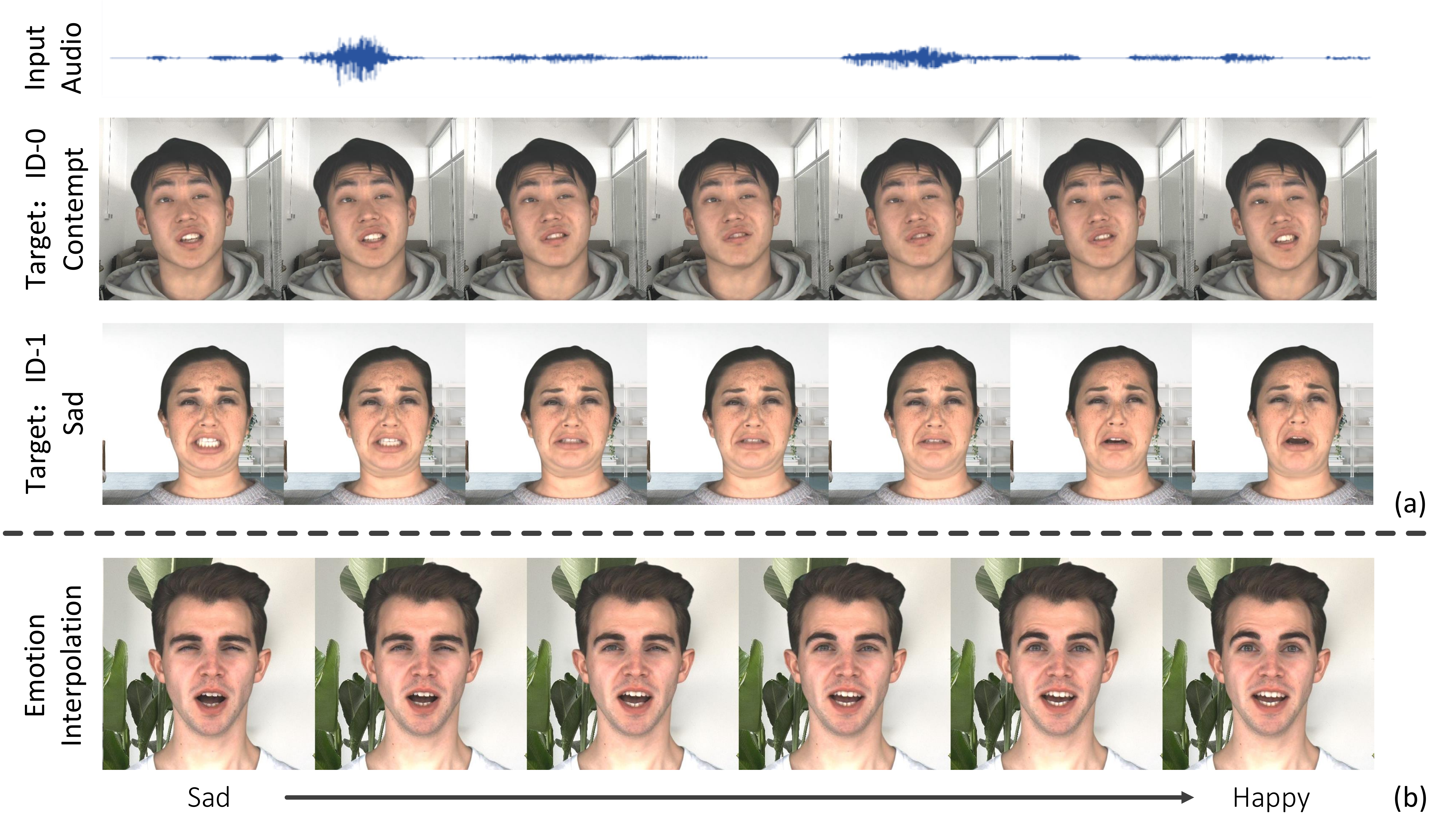}
    \captionof{figure}{\textbf{Audio-Driven Emotional Video Portraits.} Given an audio clip and a target video, our Emotional Video Portraits (EVP) approach is capable of generating emotion-controllable talking portraits and change the emotion of them smoothly by interpolating at the latent space. (a) Generated video portraits with the same speech content but different emotions (\ie, contempt and sad). (b) Linear interpolation of the learned latent representation of emotions from sad to happy.}
    \label{fig:first}
    \vspace{0pt}
\end{strip}

\begin{abstract}
Despite previous success in generating audio-driven talking heads, most of the previous studies focus on the correlation between speech content and the mouth shape.
Facial emotion, which is one of the most important features on natural human faces, is always neglected in their methods.
In this work, we present Emotional Video Portraits (EVP), a 
system for
synthesizing high-quality video portraits with vivid emotional dynamics driven by audios. 
Specifically, we propose the Cross-Reconstructed Emotion Disentanglement technique to decompose speech into two decoupled spaces, \ie, a duration-independent emotion space and a duration-dependent content space. With the disentangled features, dynamic 2D emotional facial landmarks 
can be deduced. Then we propose the Target-Adaptive Face Synthesis technique to generate the final high-quality video portraits, by bridging the gap between the deduced landmarks and the natural head poses of target videos.
Extensive experiments demonstrate the effectiveness of our method both qualitatively and quantitatively.\footnote{All materials are available at \url{https://jixinya.github.io/projects/evp/}.}

\end{abstract}

\vspace{-2mm}
\section{Introduction}
Generating audio-driven photo-realistic portrait video is of great need to multimedia applications, such as film-making~\cite{kim2019neural}, telepresence~\cite{adalgeirsson2010mebot} and digital human animation~\cite{liu2015video,edwards2016jali,zhou2018visemenet}.
Previous works have explored generating talking heads or portraits whose lip movements are synced with the input speech contents.
Generally, these techniques can be divided into two categories: 1) \textit{image-based} methods that animate one or few frames of cropped faces~\cite{chung2017you,zhou2019talking,song2018talking,chen2019hierarchical,mittal2020animating,zhou2020makeittalk,chen2020talking}, and 2) \textit{video-based editing} methods that directly edit target video clips~\cite{suwajanakorn2017synthesizing,song2020everybody,thies2019neural,wen2020photorealistic}.
Nevertheless, most of the previous studies did not model \textit{emotion}, a key factor for the naturalism of portraits.
%


Only few \textit{image-based} works have discussed emotional information in talking head generation.
Due to the lack of appropriate audio-visual datasets with emotional annotations, Vougioukas \textit{et al.}~\cite{vougioukas2019realistic} do not model emotions explicitly. 
Simply encoding emotion and audio content information into a single feature, they produce preliminary results with low quality.
%
%
Most recently, Wang \textit{et al.}~\cite{kaisiyuan2020mead} collect the MEAD dataset, which contains high-quality talking head videos with annotations of both emotion category and intensity. Then they set emotion as an one-hot condition to control the generated faces. 
However, all of these \textit{image-based} methods render only minor head movements with fixed or even no backgrounds, making them impractical in most real-world scenarios.

Whereas, \textit{video-based editing} methods, which are more applicable as discussed in ~\cite{suwajanakorn2017synthesizing,Fried2019TextbasedEO,song2020everybody,thies2019neural,wen2020photorealistic}, have not considered \textit{emotion} control. Most of them only edit the mouth and keep the upper half of the video portraits unaltered, making free emotion control unaccessible.

In this study, we propose a novel algorithm named \textbf{Emotional Video Portraits (EVP)}, aiming to endow the \textit{video-based editing} talking face generation with the ability of \textit{emotion} control from audio.  We animate full portrait 
with emotion dynamics that better matches the speech intonation, leading to more vivid results.
However, it is \textit{non-trivial} to achieve this. There exist several intricate challenges: 1) 
The extraction of emotion from audio is rather difficult, since the emotion information is stickily entangled with other factors like the speech content. 2) The blending of the edited face and the target video is difficult while synthesizing high fidelity results. Audio does not supply any cues for head poses and the global movements of a head, thus the edited head inferred from audio may have large head pose and movement variances with the target videos.

To tackle the challenges mentioned above, we manage to achieve audio-based emotion control in the proposed Emotional Video Portraits system with two key components, namely \textit{Cross-Reconstructed Emotion Disentanglement}
, and \textit{Target-Adaptive Face Synthesis}. 
%
%
To perform emotion control on the generated portraits, we firstly propose the \textit{Cross-Reconstructed Emotion Disentanglement} technique on audios to extract two separate latent spaces: i) a duration-independent space, which is a content-agnostic encoding of the emotion; ii) a duration-dependent space, which encodes the audio's speech content. Once extracted, features from these latent spaces are recombined to yield a new audio representation,
allowing a cross-reconstruction loss to be computed and optimized. 
%
However, to enable the cross-reconstructed training, paired sentences with the same content but different emotions at the same length should be provided. This is nearly unreachable in real-world scenarios. To this end, we adopt Dynamic Time Warping (DTW)~\cite{berndt1994using}, a classic algorithm in time series analysis, to help form pseudo training pairs with aligned uneven-length speech corpus.

Following previous methods~\cite{suwajanakorn2017synthesizing,chen2019hierarchical}, an audio-to-landmark animation module is then introduced with the decomposed features to deduce emotional 2D landmark dynamics.
As no pose information is provided in audio, there is a gap to be bridged between the generated landmarks and the large variances of head pose and movement in target video. 
%
To this end, we propose the \textit{Target-Adaptive Face Synthesis} technique to bridge the pose gap between the inferred landmarks and the target video portraits in 3D space. With a carefully designed 3D-aware keypoint alignment algorithm, we are able to project 2D landmarks into the target video.
Finally, we train an Edge-to-Video translation network to generate the final high-quality emotional video portraits. Extensive experiments demonstrate the superior performance of our method and the effectiveness of several key components.

\begin{figure*}[t]
\begin{center}
\includegraphics[width=1.\textwidth]{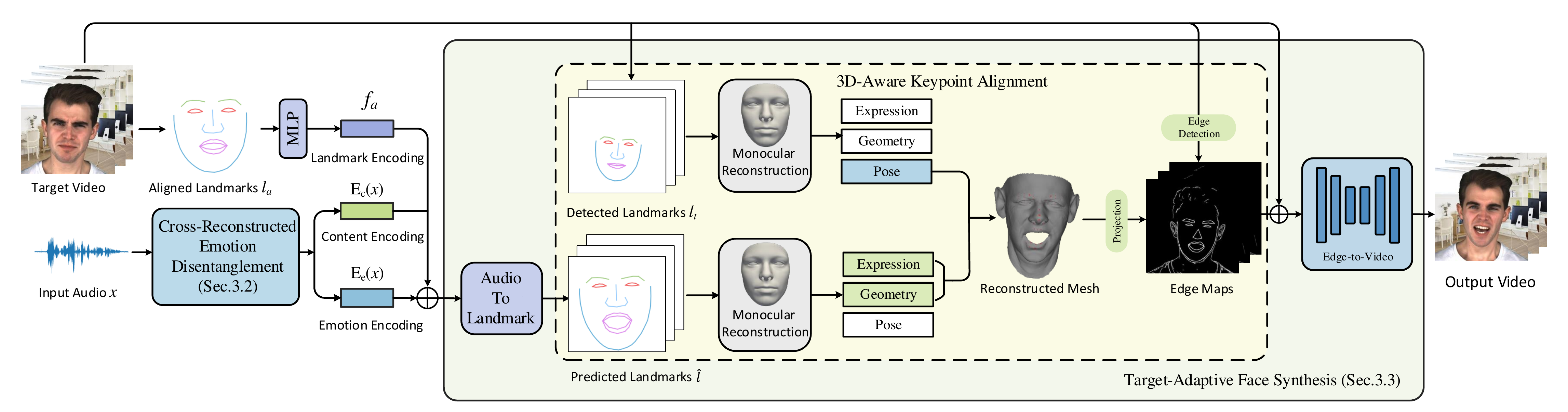}  
\end{center}
\vspace{-5mm}
\caption{\textbf{Overview of our \emph{Emotional Video Portrait} algorithm.} We first extract disentangled content and emotion information from the audio signal. Then we predict landmark motion from audio representations. The predicted motion is transferred to the edge map of the target video via a 3D-aware keypoint alignment module. Finally, the rendering network gives us photo-realistic animations of the target portrait based on the target video and edge maps.} 
\label{pipeline}
\vspace{-10pt}
\end{figure*}

Our contributions are summarized as follows:
%
%
\begin{itemize}
    \item We propose the Emotional Video Portraits (EVP) system, which is the first attempt to achieve emotional control in video-based editing talking face generation methods.
    \item We introduce Cross-Reconstructed Emotion Disentanglement technique, to distill content-agnostic emotion features for free control.
    \item We introduce Target-Adaptive Face Synthesis, to synthesize high quality portrait by making the generated face adapt to the target video with natural head poses and movements.
\end{itemize}

\section{Related Work}
\label{Related}
\subsection{Audio-Driven Talking Face Generation} 
The task of audio-driven talking-head generation aims at synthesizing lip-synced videos of a speaking person driven by audio. 
It is a topic of high demand in the field of entertainment, thus has long been the research interest in the area of computer vision and computer graphics~\cite{brand1999voice,karras2017audio,chung2017you,wang2010synthesizing,suwajanakorn2017synthesizing,zhou2019talking,song2018talking,zhu2018high,chen2018lip,chen2019hierarchical,thies2019neural,mittal2020animating,zhou2020makeittalk,kaisiyuan2020mead,chen2020comprises,Zhou2021CVPR,wen2020photorealistic}. 
%
We can divide these methods into two categories according to the differences in the visualization of their results.

\noindent\textbf{Image-Based Methods.} One type of models focuses on driving the cropped facial areas with one or more frames as the identity reference.
Chung \textit{et al.}~\cite{chung2017you}, for the first time, propose to generate lip-synced videos in an image-to-image translation~\cite{isola2017image} manner. Then Zhou \textit{et al.}~\cite{zhou2019talking} and Song \textit{et al.}~\cite{song2018talking} improve their results using disentangled audio-visual representation and recurrent neural networks. Moreover, Chen \textit{et al.}~\cite{chen2019hierarchical} leverage landmarks as intermediate representation and split the process into two stages. However, these methods can only promise the synchronization between generated mouths and audios. The results have barely any expression or head movements. Zhou \textit{et al.}~\cite{zhou2020makeittalk} successfully generate identity-related head movements, but their model also fails to control emotions. 

As for emotional talking faces, Vougioukas \textit{et al.}~\cite{vougioukas2019realistic} adopt three separated discriminators to enhance synthesis details, synchronization, and realistic expressions, respectively. However, their experiments are carried out on limited scales.
%
Most recently, Wang \textit{et al.}~\cite{kaisiyuan2020mead} propose the MEAD dataset and propose to generate emotional talking faces by splitting the manipulation for the upper and lower part of the face, respectively. Nevertheless, their results are less realistic and limited to only the facial areas.

\noindent\textbf{Video-Based Editing Methods.} Full-frame video portraits contain not only the facial areas but also the neck and the shoulder part of the person, together with the background. It is without doubt that this setting is more realistic, but more difficult to reconstruct. As a result, most methods edit only the mouth areas. Suwajanakorn \textit{et al.}\cite{suwajanakorn2017synthesizing} synthesize photo-realistic talking videos of Obama by training an audio to landmark RNN. A re-timing module is proposed for head-poses. Song \textit{et al.}\cite{song2020everybody} and Thies \textit{et al.}\cite{thies2019neural} all regress facial expression parameters of 3DMM models, and inpaint the mouth regions.
%
%
While high-quality results can be rendered through these pipelines using the videos of a target subject, it is difficult for their models to manipulate the upper face, left alone emotions. In this work, we propose to generate emotional manipulable full-frame talking-heads.

\subsection{Conditional Emotion Generation}
Inspired by the great success of unsupervised image translation\cite{CycleGAN2017, isola2017image,lee2018diverse,huang2018multimodal, choi2018stargan,jia2019gan}, several methods focusing on emotion conditioned generation have been proposed in recent years.
Ding \textit{et al.}~\cite{ding2018exprgan} design a novel encoder-decoder architecture to control expression intensity continuously by learning an expressive and compact expression code.
Pumarola \textit{et al.}~\cite{pumarola2018ganimation} introduce an unsupervised framework named GANimation, which is able to generate continuous facial expressions of a specified emotion category by activating the action units (AU) to various states. 
However, unsatisfying artifacts have always been a challenging problem for these methods due to the lack of explicit and accurate guidance.
Inspired by \cite{chen2019hierarchical}, our method also chooses facial landmarks as a more reliable intermediary to generate talking face sequences with high-fidelity emotions. 

\section{Method}
\label{method}

\subsection{Overview }  
%
%

As shown in Fig.~\ref{pipeline}, our \emph{Emotional Video Portrait (EVP)} algorithm consists of two key components. 
The first is \emph{Cross-Reconstructed Emotion Disentanglement} that learns the disentangled content and emotion information from the audio signals. 
We use a temporal alignment algorithm, Dynamic Time Warping~\cite{berndt1994using} to produce pseudo training pairs, and then design a Cross-Reconstructed Loss for learning the disentanglement (Sec.~\ref{sec:caed}). 
%
The second part of our algorithm is \emph{Target-Adaptive Face Synthesis} that adapts the facial landmarks inferred from the audio representations to the target video. 
 %
We design a 3D-Aware Keypoint Alignment algorithm to rotate landmarks in 3D space, and thus the landmarks can be adaptive to various poses and motions in the target video.
%
%
Then, we subsequently use the the projected 2D landmarks as the guidance to edit the target video via an Edge-to-Video network(Sec.~\ref{sec:tafs}).
In the following sections, we describe each module of our algorithm in detail.

\begin{figure}[t]
\begin{center}
\includegraphics[width=.45\textwidth]{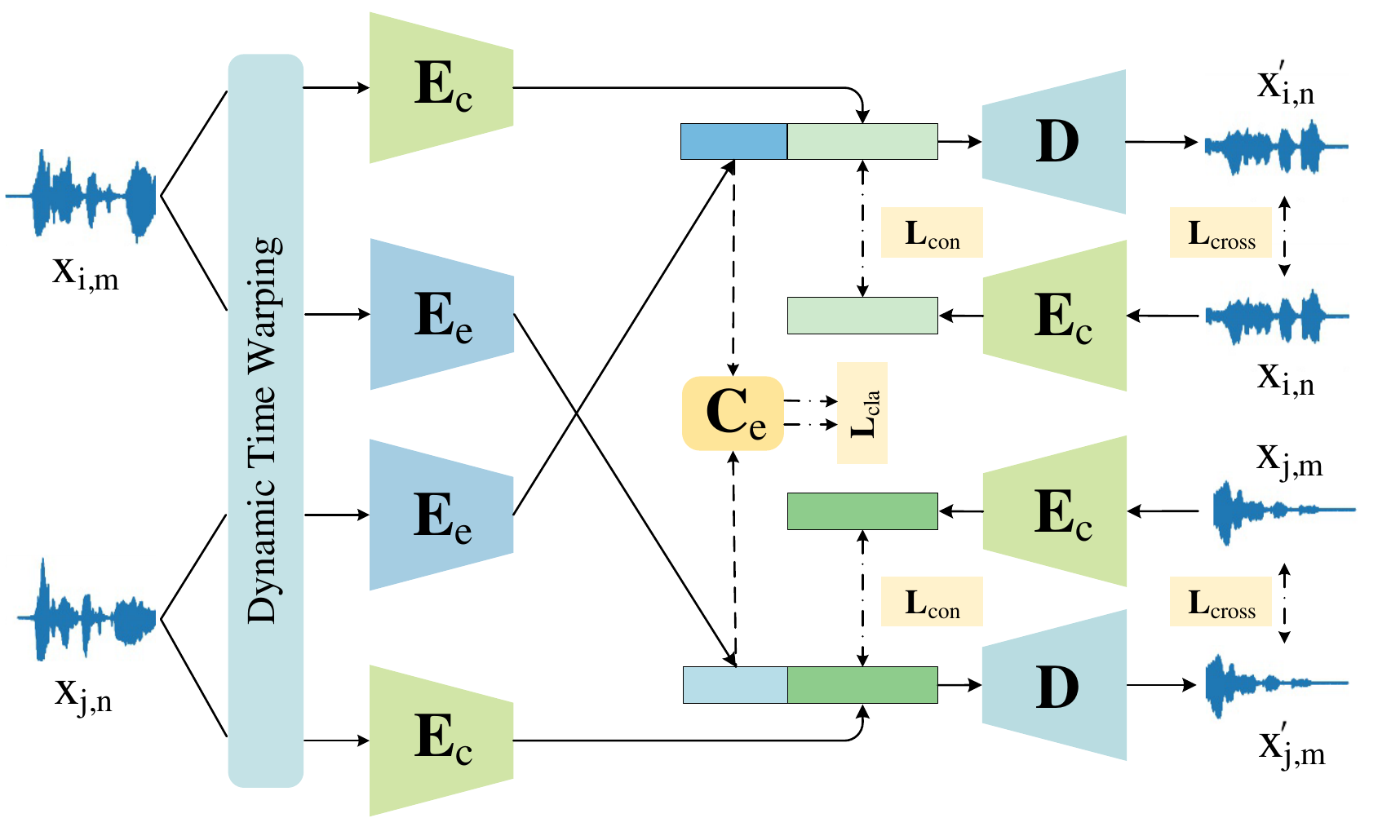} 
\end{center}
\vspace{-3mm}
\caption{\textbf{Cross-reconstruction for disentanglement.} The emotion and content representations extracted from different audio signals are combined to reconstruct corresponding samples. Part of the training losses are also shown on this figure.}
\vspace{-0.4cm}
\label{disentangle}
\end{figure}

\subsection{Cross-Reconstructed Emotion Disentanglement }
\label{sec:caed}

To achieve audio-based emotional control for talking face synthesis, the inherently entangled emotion and content components need to be independently extracted from audio signals.
%
%
Unlike previous methods~\cite{kaisiyuan2020mead}, which learn one single representation from audio signals, we propose to extract two separate latent audio spaces: i) a duration-independent space,  which  is  a  content-agnostic  encoding of the emotion and  ii) a duration-dependent space, which encodes the audio’s speech content.
While cross-reconstruction~\cite{aberman2019learning} technique seems promising on such a task, it  can only be enabled through paired audio clips with the same content but different emotions at the same length. Nevertheless, this is nearly unreachable in real-world scenarios. To this end, we firstly build aligned \textit{pseudo training pairs}, and then adopt the \textit{cross-reconstructed training} for emotion disentanglement in audios.

\noindent{\textbf{Build Pseudo Training Pairs.}}
An audio-visual dataset~\cite{kaisiyuan2020mead} with various characters speaking the same corpus under different emotion states is leveraged to train this disentanglement network.
Since the speeches with the same content but different emotions vary in speech rate, 
we resort to a temporal alignment algorithm to align the uneven-length speeches.

Specifically, we use Mel Frequency Cepstral Coefficients (MFCC) \cite{logan2000mel} as audio representations and use the Dynamic Timing Warping (DTW)~\cite{berndt1994using} algorithm to warp the MFCC feature vectors by stretching or shrinking them along the time dimension.
%
%
Given two MFCC sequences $\bm{S}_a$ and $\bm{S}_b$ of the same content but different lengths, DTW calculates a set of index coordinate pairs $\left \{ (i,j),...\right \}$ by dynamic programming to force $\bm{S}_a[i]$ and $\bm{S}_b[j]$ to be similar.
The optimal match between the given sequences is achieved by minimizing the sum of a distance cost between the aligned MFCC features:
\begin{equation}
\begin{aligned}
\min \sum_{(i,j)\in P} d(\bm{S}_a[i],\bm{S}_b[j]),
\end{aligned}
\end{equation}
where $d$ is the distance cost, $P$ is the path for alignment. The path constraint is that, at $(i,j)$, the valid steps are $(i+1,j)$, $(i,j+1)$, and $(i+1,j+1)$, making sure that the alignment always moves forward each time for at least one of the signals.
%
%
These aligned audio samples can then be used as the inputs to the disentanglement network for cross-reconstruction.

\noindent{\textbf{Cross-Reconstructed Training.}} The cross-reconstructed training procedure is as shown in Fig.~\ref{disentangle}.
To independently extract the emotion and content information lie in an audio clip $x_{i, m} $ with content $i$ and emotion $m$, two encoders $E_c$ and $E_e$ are leveraged for embedding the two information respectively. Intuitively, when the two representations are completely disentangled, we can use the information in both the content embedding $E_c(x_{i, m})$ and the emotion embedding $E_c(x_{j, n})$ from audio clips $x_{i, m}$ and $x_{j, n}$ to reconstruct the clip $x_{i, n}$ from a decoder $D$. By leveraging the pseudo training pairs we build before, we introduce two new samples $x_{i,n}, x_{j,m}$ to serve as supervisions for the reconstruction procedure.
Since each sample can only provide one type of information that is beneficial to the cross-reconstruction, the disentanglement can be finally achieved.

We supervise the training process with a loss function including four parts: cross reconstruction loss, self reconstruction loss, classification loss, and content loss.
%
%
Given four audio samples $x_{i,m},x_{j,n},x_{j,m},x_{i,n}$, we formulate the cross reconstruction loss as :
\begin{equation}
\begin{aligned}
 L_{cross} &= \| D(E_c({x}_{i,m}),E_e({x}_{j,n})) - {x}_{i,n} \|_2 \\
&+\| D(E_c({x}_{j,n}),E_e({x}_{i,m})) - {x}_{j,m} \|_2.
\end{aligned}
\end{equation}

Besides, we are also able to reconstruct the original input by using the encoders and the decoder, namely the self reconstruction loss defined as:
\begin{equation}
\begin{aligned}
 L_{self} &= \| D(E_c({x}_{i,m}),E_e({x}_{i,m})) - {x}_{i,m} \|_2 \\
&+\| D(E_c({x}_{j,n}),E_e({x}_{j,n})) - {x}_{j,n} \|_2. 
\end{aligned}
\end{equation}

In order to encourage the $E_e$ to map samples with the same emotion type into clustered groups in the latent space, we add a classifier $C_e$ for the emotion embedding and an additional classification loss defined as:
\begin{align}
 L_{cla} = -\sum_{k=1}^{N}(p_{k}\ast\log q_{k}). 
\end{align}
Here, $N$ denotes the number of different emotion types, $p_{k}$ denotes whether the sample takes emotional label $k$, and $q_k$ denotes the corresponding network prediction probability.
Moreover, we also constrain the samples with the same utterance to share similar content embedding:
\begin{align}
 L_{con} = \sum_{k=i,j}\|E_c({x}_{k,m}) - E_c({x}_{k,n}) \|_1. 
\end{align}
Summing these four terms, we obtain the total loss function:
\begin{align}
L_{dis} = L_{cross} + L_{self} + \lambda_{cla}L_{cla} + \lambda_{con}L_{con}, 
\end{align}
where $\lambda_{cla}$ and $\lambda_{con}$ are weights for the classification and the content loss respectively.

\subsection{Target-Adaptive Face Synthesis }
\label{sec:tafs}
To generate photo-realistic facial animations of the input portrait,
we first introduce an \emph{audio-to-landmark} network, following \cite{suwajanakorn2017synthesizing,chen2019hierarchical}, that predicts landmark motions from the disentangled audio embeddings.
Afterwards,
video-based editing methods normally perform facial editing on a target video clip recorded on the target person. However, this would lead to two challenges in our setting which alter the whole face rather than only the mouth: 1) The misalignment of head poses. The head movements of the predicted landmarks may differ from the target video severely, and barely any pose information is provided in the audio. 2) The blending of the edited face and the target video is difficult while synthesizing high fidelity results. 
 
To cope with such challenges, we propose the \textit{3D-Aware Keypoint Alignment} algorithm to align our generated landmarks with guidance landmarks in the 3D space. Then we propose to merge our generated landmarks with the edge map of the target image. The two of them together can serve as the guidance to train an \emph{Edge-to-Video} translation network for the final results.


\noindent
\textbf{Audio-to-Landmark Module.}
Our first goal is to learn landmark displacements from emotional audio clips, thus requiring the facial shape, or identity information from the aligned landmark $l_a$ unchanged. So we extract the landmark identity embedding $f_a$ with a multi-layer perceptron (MLP) as shown in Fig~\ref{pipeline}.
%
%
%
Then $f_a$ is sent into an audio-to-landmark module along with the two disentangled audio  embeddings $E_c(x)$ and $E_e(x)$. The audio-to-landmark module predicts the landmark displacements $l_d$ by a long short-term memory (LSTM) network followed by a two-layer MLP.

In terms of the loss function, we minimize the distance between the reference landmarks $l$ and the predicted ones $\hat{l}$ defined below:
%
\begin{equation}
\begin{aligned}
L_{a} &= \| \hat{l} - l \|_2 = \| l_a + l_d - l \|_2.
\end{aligned}
\end{equation}
%
%

\noindent
\textbf{3D-Aware Keypoint Alignment.}
For aligning head poses, we first perform a landmark detection on the target video using an off-the-shelf method~\cite{Wu_2018_CVPR}. Then we operate in the 3D space, where the pose information is explicitly defined.
A parametric 3D face model~\cite{cao2013facewarehouse} is utilized to recover the 3D parameters from 2D landmarks by solving a non-linear optimization problem.
%
%
Utilizing the 3D geometry and expression parameters, we get a set of pose-invariant 3D landmarks $\bm{L}^{3d}_p$ (See supplementary for details).
The pose parameters $\bm{p}$ contains a $3 \times 3$ rotation matrix $\bm{R}$, 2 translation coefficients $\bm{t}$, and 1 scaling coefficient $s$. 
%
%
%
By replacing the pose parameters of the predicted landmark with the detected ones in the target video ($\bm{R}_t,\bm{t}_{t}, s_t$), we obtain the adapted 3D key points and then project them to the image plane with scale orthographic projection:
\begin{equation}
\begin{aligned}
\bm{l_p} = s_t * \bm{Pr} * \bm{R}_t * \bm{L}^{3d}_p + \bm{t}_{t},
\end{aligned}
\end{equation}
where $\bm{l_p}$ is the projected 2D landmark and $\bm{Pr}$ is the orthographic projection matrix $\begin{pmatrix}
1 & 0 & 0 \\
0 & 1 & 0
\end{pmatrix}$.
%
%
Since the geometry and expression parameters remain unchanged, the projected landmarks $\bm{l_p}$ naturally share consistent identity and facial expressions with the predicted landmarks.
While the head pose, scale and position are set the same as the face in the target video frame.

Note we use 3D models only for pose alignment. An alternative way is to directly predict 3D facial parameters for facial reconstruction~\cite{kim2018deep,Zhou_2020_CVPR}. However,  inaccurate fitted facial expressions are not sufficient for representing the detailed emotions in our work.


\begin{figure*}[t]
\begin{center}
\setlength{\abovecaptionskip}{0.cm}
\includegraphics[width=1.\textwidth]{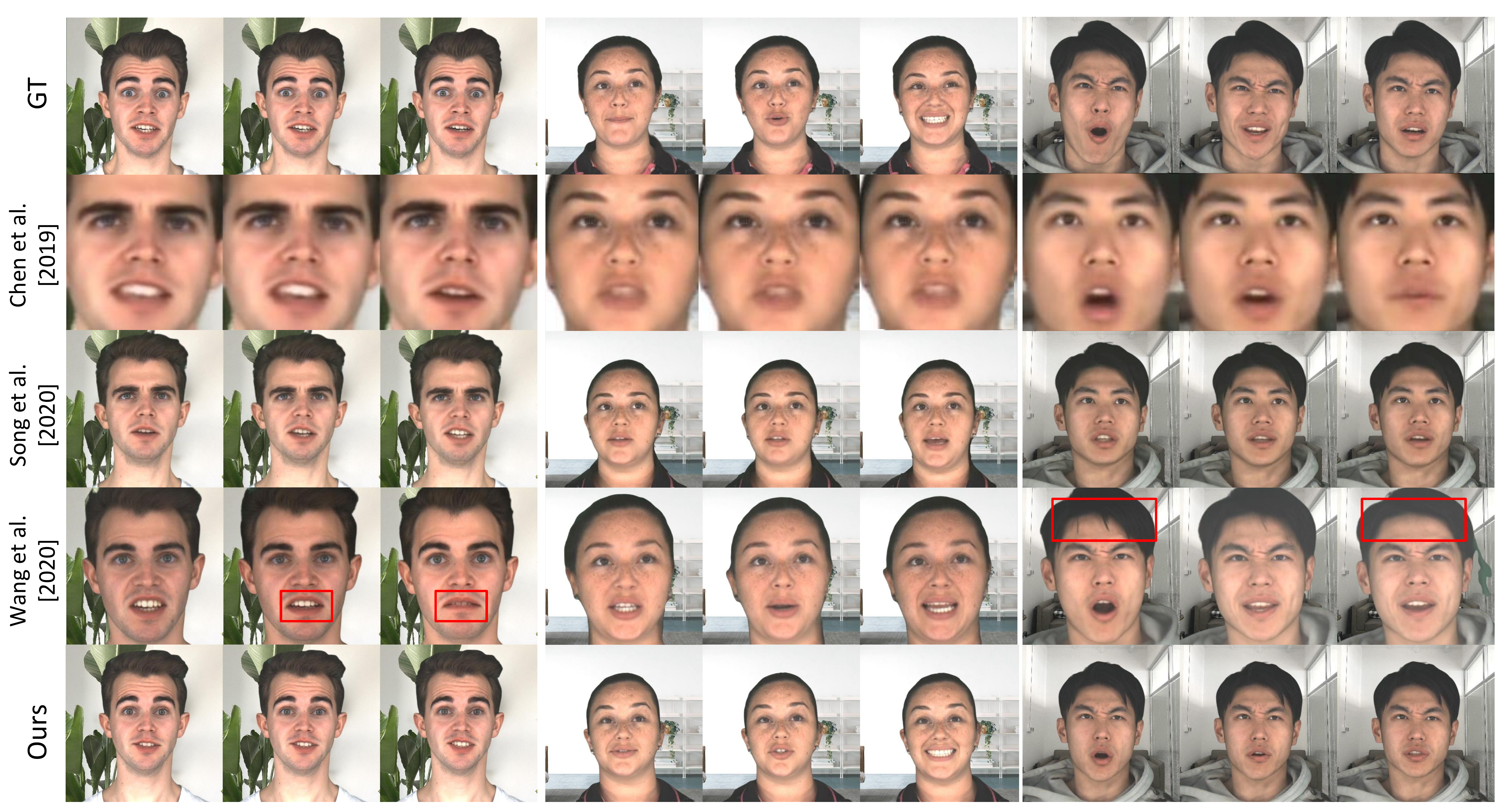} 
\end{center}
\vspace{-0.4cm}
\caption{\textbf{Qualitative comparisons with the state-of-the-art methods.} We show three examples with different speech content and emotions. Note that we choose the same target video with frontal face for Song \textit{et al.}\cite{song2020everybody} and ours, and we use the first frame of the target video for Chen \textit{et al.}\cite{chen2019hierarchical} and Wang \textit{et al.}\cite{kaisiyuan2020mead} as they edit a target image rather than a target video.} 
\label{img1}
\end{figure*}

\setlength{\tabcolsep}{12pt}
\begin{table*}[t!]
\begin{center}
\begin{tabular}{c|c|c|c|c|c|c|c}
\toprule
Method/Score& M-LD $\downarrow$ & M-LVD $\downarrow$ & F-LD $\downarrow$ & F-LVD $\downarrow$ & SSIM $\uparrow$ & PSNR $\uparrow$ & FID $\downarrow$\\
\hline 
Chen \textit{et al.}\cite{chen2019hierarchical}          &3.27    &2.09   &3.82    &1.71 &0.60    &28.55   &67.60\\
Wang \textit{et al.}\cite{kaisiyuan2020mead}             &2.52    &2.28   &3.16    &2.01 &0.68    &28.61   &22.52\\
Song \textit{et al.}\cite{song2020everybody}              &2.54   &1.99   &3.49    &1.76   &0.64    &29.11   &36.33\\
Ours             &\textbf{2.45}   &\textbf{1.78}   &\textbf{3.01}    &\textbf{1.56}  &\textbf{0.71}   &\textbf{29.53}   &\textbf{7.99}\\
\bottomrule
\end{tabular}
\vspace{-0.2cm}
\caption{\small{\textbf{Quantitative comparisons with the state-of-the-art methods.} We calculate the landmark accuracies and video qualities of the results of different solutions by comparing them with the ground truth. M- represents mouth and F- stands for face region.} }
\vspace{-0.4cm}
\label{metric}
\vspace{-3mm}
\end{center}
\end{table*}

\noindent
\textbf{Edge-to-Video Translation Network.}

Given the adapted landmarks and the target frame,  we merge the landmarks and the edge map extracted from this frame into a guidance map for portrait generation.
%
In particular, we extract edges outside the face region using an edge detection algorithm~\cite{green2002canny}, and replace the original landmarks with our aligned ones. Then we connect adjacent facial landmarks to create a face sketch.
%

Following~\cite{wang2018vid2vid}, we adopt a conditional-GAN architecture for our Edge-to-Video translation network.
%
The generator part $G$ is designed as a coarse-to-fine architecture~\cite{wang2018high}, 
%
while the discriminator part is designed to guarantee both the quality and the continuity of the generated frames.
Please refer to \cite{wang2018vid2vid} for more details about the network architecture.
%
%
%
%
%

\section{Experiment}
\noindent\textbf{Implementation Details.}
We evaluate our method on MEAD, a high-quality emotional audio-visual dataset with 60 actors/actresses and eight emotion categories. The models are trained and tested on the train/test splits of the dataset.
All the emotional talking face videos are converted to 25 fps and the audio sample rate is set to be 16kHz. 
For the video stream, we align all the faces based on the detected facial landmarks.
As for the audio stream, we follow the design in \cite{chen2019hierarchical} to extract a 28$\times$12 dim MFCC feature corresponding to each frame in the video. 
Before training the disentanglement module, the emotion encoder is pretrained through an emotion classification task \cite{ooi2014new}.
%
%
Meanwhile, the content encoder is pretrained on LRW \cite{chung2016lip}, a lip-reading dataset with barely any emotion.
Then we discard the decoder and use the two pretrained encoders in our training process.
More implementation details can be found in the supplementary materials.
%
%
%

\noindent\textbf{Comparing Methods.}
We compare our work with three prior works \cite{chen2019hierarchical,kaisiyuan2020mead,song2020everybody}. 
%
%
The method of Chen \textit{et al.}~\cite{chen2019hierarchical} is an image-based method, which synthesizes facial motions based on landmarks and employs an attention mechanism to improve the generation quality.
The method of Song \textit{et al.}~\cite{song2020everybody} is a video-based method which applies 3D face models to realize audio-based video portrait editing. 
Then we compare our work with Wang \textit{et al.}~\cite{kaisiyuan2020mead}, the most relevant work that proposes the first talking face generation approach with the capacity of manipulating emotions.
We believe these three works are the most representative works to compare with. 


\begin{figure}[t]
\begin{center}
\includegraphics[width=1.\linewidth]{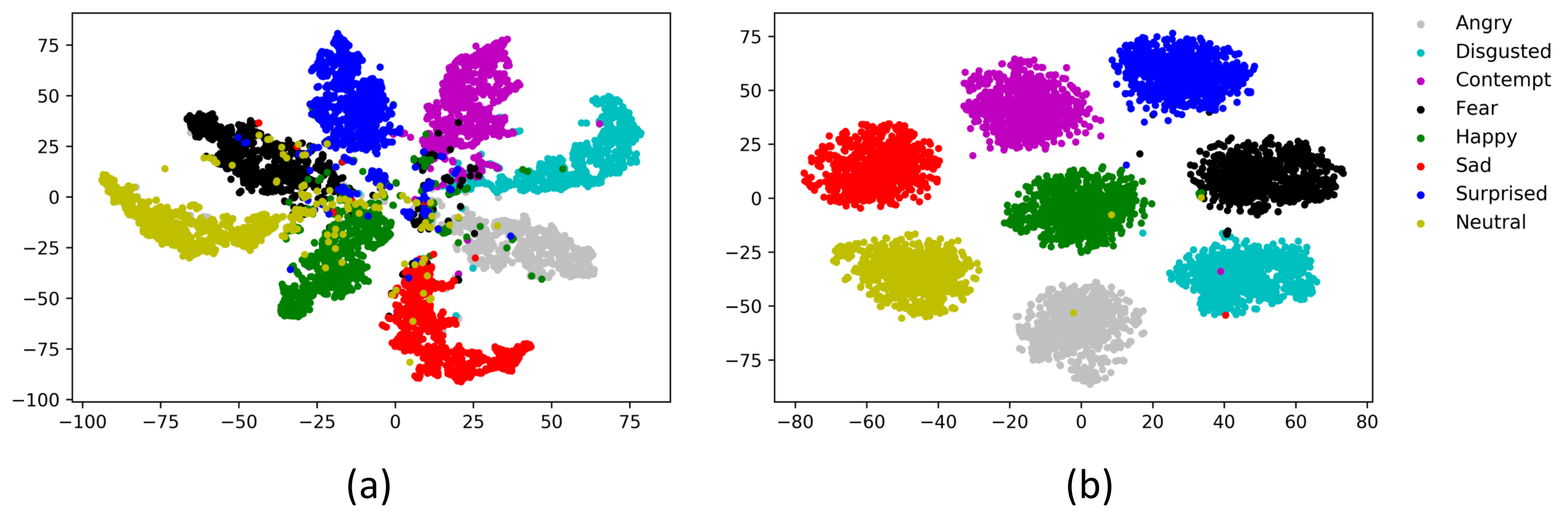} 
\end{center}
\vspace{-4mm}
\caption{\textbf{Emotion latent space clusters with and without the cross-reconstruction part.} (a) Emotion latent codes of the pretrained emotion encoder. (b) Emotion latent codes of the encoder trained by cross-reconstruction. Different colors indicate different emotion types. }
\label{img4}
\vspace{-2mm}
\end{figure}


\subsection{Experimental Results }
\noindent\textbf{Qualitative Comparisons.}
We make comparisons with other methods on various sequences as shown in the accompanying video. We also select some frames as shown in Fig.~\ref{img1}.
%
%
Our method is able to generate high-fidelity emotional talking face video which is better than others.
Concretely, Chen \textit{et al.}\cite{chen2019hierarchical} and Song \textit{et al.}\cite{song2020everybody} do not consider emotions, so they generate plausible mouth shapes but always with the neutral emotion.
%
%
Wang \textit{et al.}\cite{kaisiyuan2020mead} is able to generate desired emotions. 
However, the emotion of the predicted mouth shape is sometimes inconsistent with the facial expression (left) since it directly learns mouth shapes from audio signals where the emotion and content information are closely entangled. 
In addition, the algorithm in Wang \textit{et al.}\cite{kaisiyuan2020mead} is not robust enough to data with large head movements and background variations, leading to implausible facial expressions (middle) and changes with characteristics like hairstyles (right).
%
%

\noindent\textbf{Quantitative Comparisons.}
To quantitatively evaluate different methods, we extract facial landmarks from the aligned result sequences and the ground truth sequences. The alignment is also for compensating head motions.
Then, the metrics of Landmark Distance(LD) and Landmark Velocity Difference(LVD)\cite{chen2019hierarchical,zhou2020makeittalk} are utilized to evaluate facial motions.
LD represents the average Euclidean distance between generated and recorded landmarks. 
Velocity means the difference of landmark locations between consecutive frames, so LVD represents the average velocity differences of landmark motions between two sequences. 
We adopt LD and LVD on the mouth and face area to evaluate how well the synthesized video represents accurate lip movements and facial expressions separately. 
%
%
To further evaluate the quality of the generated images of different methods, we compare the SSIM~\cite{wang2004image}, PSNR, and FID~\cite{heusel2017gans} scores. The Results are illustrated in Table~\ref{metric}. 
Our method obviously outperforms others in audio-visual synchronization (M-LD, M-LVD), facial expressions (F-LD, F-LVD) and video quality (SSIM, PSNR, FID). 
%

\begin{figure}[t]
\begin{center}
\includegraphics[width=1.\linewidth]{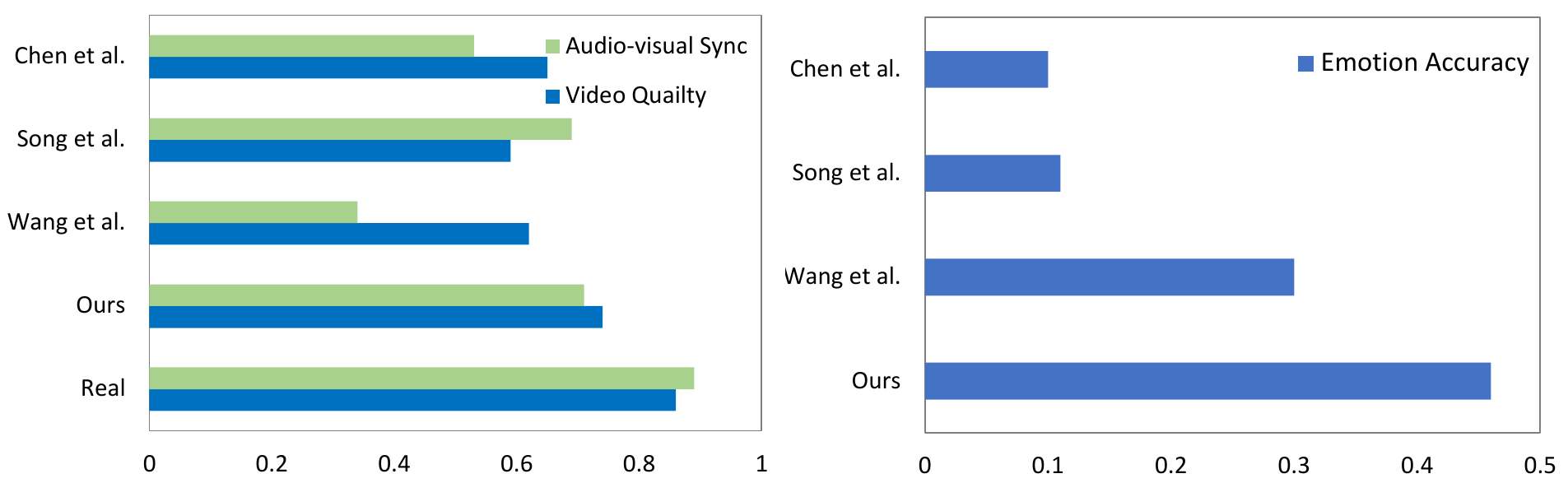} 
\end{center}
\caption{\textbf{User study.} User study results of audio-visual synchronization, video quality and emotion accuracy.}
\label{User-study}
\vspace{-4mm}
\end{figure}

\noindent\textbf{User Study.}
To quantify the quality (including the accuracy of emotion and facial motion) of the synthesized video clips, we design thoughtful user studies to compare real data with generated ones from EVP, Wang \textit{et al.}\cite{kaisiyuan2020mead} , Chen \textit{et al.}\cite{chen2019hierarchical} and Song \textit{et al.} \cite{song2020everybody} .
We generate 3 video clips for each of the 8 emotion categories and each of the 3 speakers, hence 72 videos in total.
They are evaluated w.r.t three different criteria: whether the synthesized talking face video is realistic, whether the face motion sync with the speech, and the accuracy of the generated facial emotion.
The evaluation consists of two stages. First, the attendees are asked to judge the given video upon audio-visual synchronization and video quality and score from 1 (worst) to 5 (best).
Then we show them real emotional video clips without background sound. After that they need to choose the emotion category for the generated video without voice.
%
50 participants finished our questionnaire and the results are shown in Figure~\ref{User-study}.
As can be seen, our method obtains the highest score on visual quality and audio-visual sync apart from the real data.
We also achieve the highest accuracy on emotion classification compared with other methods.

\noindent\textbf{More Results.}
We show image results of our EVP algorithm in Figure~\ref{fig:first} and more results can be found in our supplementary video\footnote{The materials are available at \url{https://jixinya.github.io/projects/evp/}.}. 
Our method can synthesize high-quality emotional video portraits adaptive to various head poses and backgrounds.
What's more, during inference different audio signals or even learned features can be taken as content and emotion encoder inputs, leading us to more applications described in Sec.~\ref{sec:disen}.


\subsection{Disentanglement Analysis }
\label{sec:disen}

As illustrated in Sec.~\ref{method}, our network disentangles content and emotion information from audio signals.
To validate this, we feed different audio inputs to the content encoder and the emotion encoder. 
%
As shown in part (a) of Fig.~\ref{fig:first}, the mouth motion in the generated video is in accordance with the audio fed into the content encoder, while the generated facial expression matches the emotion of the audio of the emotion encoder.
%
%
Extensive experiments (shown in the accompanying video) also indicate that the speech content and emotion are successfully decoupled from the audio signals. 

Moreover, to quantitatively evaluate the generated emotions in the final video, we adopt an off-the-shelf emotion classification network \cite{meng2019frame} in our experiments.
We trained the classification network on the training set of MEAD and calculated the numerical emotion accuracy of a face video by comparing its emotion classification result with the ground truth label.  
Since the videos in MEAD have ground truth emotion labels, quantitative evaluation can be performed here.
%
The testing set of MEAD gets 90.2\% accuracy, indicating the classification network outputs reasonable emotion labels.
%
Our method gets 83.58\% accuracy exceeding the 76.00\% accuracy of Wang \textit{et al.}\cite{kaisiyuan2020mead}, showing that our method better maintains the emotion. 


\noindent\textbf{Emotion Editing.}  As our method encodes emotion features in a continuous latent space, alternating features in the latent space is able to achieve emotion manipulation including emotion category as well as the intensity manipulation .
%
%
In particular, we perform emotion category manipulation by calculating the mean latent feature of each emotion cluster and interpolating between the mean codes. 
%
Results are shown in part (b) of Fig.~\ref{fig:first}. By tuning the weight $\alpha$ between the source emotion $\text{E}_s$ and the target emotion $\text{E}_t$, we get image sequences conditioned on a linear interpolated emotion feature: $\alpha\text{E}_s+(1-\alpha)\text{E}_t$.
%
%
We can find that the emotion transformation between frames is consistent and smooth, which means our work is capable of continuously editing emotion via speech features.

\begin{table}[t] \footnotesize

 \vspace{-1pt}
\setlength{\tabcolsep}{10pt}
\begin{center}
\begin{tabular}{c|c|c|c|c} 
\toprule
Method/Score &M-LD  & M-LVD  & F-LD  & F-LVD \\
\hline 
Ours w/o $L_{cla}$          &2.72    &1.83   &3.68 &1.63\\
Ours w/o $L_{con}$          &2.65    &1.86   &3.03  &1.60\\ 
Ours w/o $L_{self}$         &2.47    &1.83  &3.02  &1.62\\
Ours w/o $L_{cross}$          &2.54    &1.80  &3.19 &1.59\\
Ours           &\textbf{2.45}      &\textbf{1.78}   &\textbf{3.01}  &\textbf{1.56}\\
\bottomrule
\end{tabular}
\caption{\small{\textbf{Quantitative ablation study for Cross-Reconstructed Emotion Disentanglement component.} We show quantitative results of landmarks with different losses. } }
\label{Ablation_study}
\end{center}

 \vspace{-15pt}
\end{table}

\subsection{Ablation Study }


\noindent\textbf{Cross-Reconstructed Emotion Disentanglement.} The cross-reconstruction is the key to our disentanglement. To evaluate the disentanglement, in Fig.~\ref{img4}, we compare the emotion latent spaces obtained by networks with and without the cross-reconstruction training. We use t-SNE~\cite{maaten2008visualizing} to visualize the latent codes. 
Different colors represent audio with eight different emotion categories. 
It can be seen that by using the cross-reconstruction, the samples with the same emotion class are more clustered than those obtained without it. 
This indicates that the cross-reconstruction does contribute to decoupling emotion information from audios.
%
We also evaluate the effectiveness of our cross-reconstruction by comparing the emotion classification accuracy of the final synthesized video clips. 
Without the reconstruction part, our method gets an accuracy of 69.79\%, lower than results with it(83.58\%).
It demonstrates that the reconstruction module enhances the emotion correctness of the final talking head videos and facilitates the emotion control of our technique.
Moreover, we conduct experiments to demonstrate the contributions of the four losses introduced in Sec.~\ref{sec:caed}. 
Quantitative results are shown in Table~\ref{Ablation_study}, which prove that each loss contributes to the component.

\begin{figure}[t]
\begin{center}
\includegraphics[width=1.\linewidth]{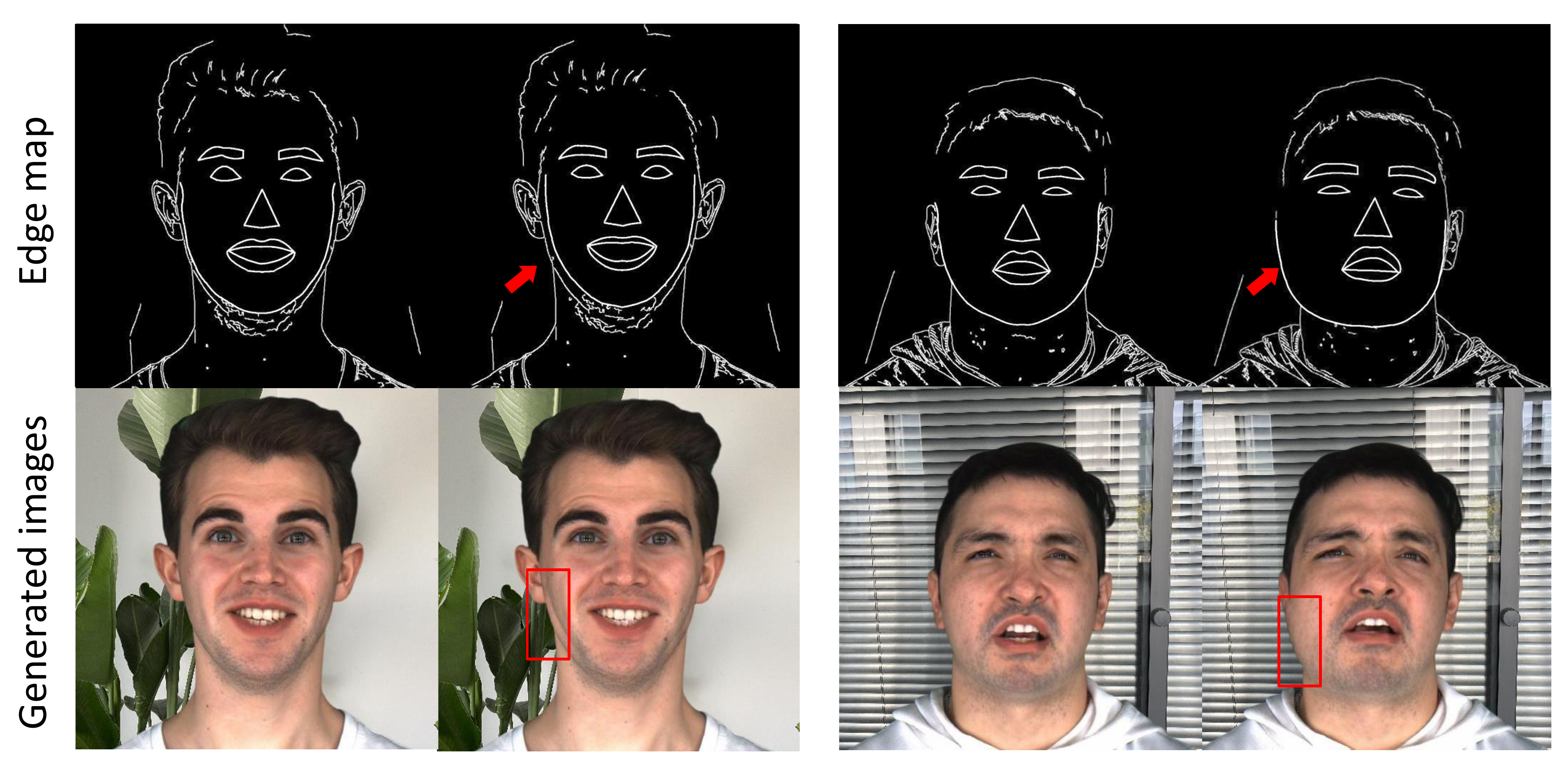}
\end{center}
\vspace{-3mm}
\caption{\textbf{Ablation study for 3D-aware keypoint alignment module.} We show cases with (left) and without (right) 3D keypoint alignment. The red arrows point out the displacements between landmarks and the face contour, while the red boxes show the artifacts in the synthesized frame.}
\label{img5}
\vspace{-3mm}
\end{figure}

\noindent\textbf{Target-Adaptive Face Synthesis.} Fig.~\ref{img5} shows the qualitative results of the 3D alignment in face synthesis.
3D landmark alignment enables us to change head motions to be consistent with the target portrait video, so that the Edge-to-Video Translation network generates smooth and realistic frames. 
On the other hand, directly using 2D facial landmarks brings displacements between the synthesized face and the face contour of the input video in the edge map, which results in noticeable artifacts in the final synthesized video. 

\section{Conclusion}
In this paper, we present an audio-driven video editing algorithm to synthesize emotional video portraits via effective learning in the decoupled representation space.
%
We propose  \emph{Cross-Reconstructed Emotion Disentanglement} to decompose the input audio sample into a pair of disentangled content and emotion embeddings, based on which, 2D facial landmarks with emotion dynamic can be generated. Then, we propose  \emph{Target-Adaptive Face Synthesis} to produce the video portraits with high fidelity by aligning the newly generated facial landmarks with the natural head poses of target videos.
Qualitative and quantitative experiments have validated the effectiveness of our method. \\

{\noindent\textbf{Acknowledgements.} This work is  supported partly by the NSFC (No.62025108), the National Key R\&D Program of China 2018YFA0704000, partly by the Beijing Natural Science Foundation (JQ19015), the NSFC (No.61822111, 61727808, 61627804), the NSFJS (BK20192003), partly by Leading Technology of Jiangsu Basic Research Plan under Grant BK2019200, and partly by A*STAR through the Industry Alignment Fund - Industry Collaboration Projects Grant.}

\newpage
{\small
\bibliographystyle{ieee_fullname}
\bibliography{egbib}
}

\end{document}